# Prediction of Herd Life in Dairy Cows Using Multi-Head Attention Transformers

Mahdi Saki[a], Justin Lipman[a]

[a] School of Electrical and Data Engineering, Faculty of Engineering and IT, University of Technology Sydney, Ultimo, NSW 2007, Australia, , mahdi.saki@uts.edu.au, justin.lipman@uts.edu.au

**Abstract**
Dairy farmers should decide to keep or cull a cow based on an objective assessment of her likely performance in the herd. For this purpose, farmers need to identify more resilient cows, which can cope better with farm conditions and complete more lactations. This decision-making process is inherently complex, with significant environmental and economic implications. In this study, we develop an AI-driven model to predict cow longevity using historical multivariate time-series data recorded from birth. Leveraging advanced AI techniques, specifically Multi-Head Attention Transformers, we analysed approximately 780,000 records from 19,000 unique cows across 7 farms in Australia. The results demonstrate that our model achieves an overall determination coefficient of 83% in predicting herd life across the studied farms, highlighting its potential for practical application in dairy herd management.

## 1   Introduction

The natural lifespan of dairy cows is about 20 years (De Vries & Marcondes, 2020). However, dairy cows leave herds much earlier within 4-6 years based on farmers' decisions (De Vries & Marcondes, 2020). Dairy farmers decide to cull or keep a cow based on animal health and farm management practices (Owusu-Sekyere et al., 2023). This is a complicated decision (Owusu-Sekyere et al., 2023) and an early prediction of animal lifespan can help dairy farmers to improve their management and breeding decisions (Ouweltjes et al., 2021), (Ranzato et al., 2022), improve animal welfare (Mulder & Rashidi, 2017) and increase farm productivity (Colditz & Hine, 2016), reduce environmental effects (Herzog et al., 2018) and reduce antibiotic usage (König & May, 2019). In other words, for a sustainable and optimal management system with respect to lifespan, farmers need to identify cows that are more resilient. This can help farmers to avoid early culling as resilient cows can cope well with farm conditions and complete several lactations (Adriaens et al., 2020).

With advances in farm management technologies in last decades, most farmers have been able to collect and record various data over animal lifetime. This has motivated researchers to apply data-driven methods for several farm management tasks like treatment (E. de Jong et al., 2023), (Ellen de Jong et al., 2023), (Bates et al., 2020) and culling decisions (Ouweltjes et al., 2021), (Ranzato et al., 2022), (Kharitonov et al., 2022), (Honnette et al., 1980), (VanRaden & Klaaskate, 1993). Although, early predictions of lifespan can help farmers to make optimal decisions for culling, it can also improve treatment decisions as farmers can identify animals that are or are not worth treating with antibiotics (Adriaens et al., 2020).

Two different types of lifespan can be defined for dairy cows from the literature (Schuster et al., 2020): herd life (HL), which is defined as the time from birth to culling dates; and production life (PL), which is the time between the first calving and culling dates. Therefore, by estimating HL for a cow, her PL can be also adjusted from her age at first calving.

Several works used historical data to build models that can estimate lifespan in dairy cows (Adriaens et al., 2020), (Ouweltjes et al., 2021), (Ranzato et al., 2022), (Honnette et al., 1980), (VanRaden & Klaaskate, 1993), (Cruickshank et al., 2002). The difference between those works is based on two aspects including the data used for building models as well as their modelling approaches. Table I shows such differences from both the aspects. The differences in data-related factors include attributes used, breeds considered, animal age ranges, total records, number of unique cows and the diversity of farms from which the data was collected. Regarding the modelling approaches, 2 main approaches can be summarized including linear and nonlinear models. More details have been provided as follows.

The earliest works used data-driven approaches to predict lifetime in dairy cows are included (Honnette et al., 1980), (VanRaden & Klaaskate, 1993) and (Cruickshank et al., 2002). Authors in (Honnette et al., 1980) developed multiple linear models, with the most effective model achieving a determination coefficient ($R^2$) of up to 11.2% for HL predictions. (VanRaden & Klaaskate, 1993) utilized multiple variables, including cumulative and current months in milk (MIM), age at first lactation, current months dry, and lactation status, to predict the productive life of dairy cows. By applying a linear regression approach, they achieved $R^2$ values ranging from 8% to 59% for cows aged 36 to 72 MIM, respectively. (Cruickshank et al., 2002) developed a linear model to predict HL based on a combination of production and genetic traits for cows with the potential to reach 84 months of age. Using linear type traits alone, their model achieved an $R^2$ of up to 50% for HL predictions. A second model, incorporating both yield and type traits, demonstrated improved performance with an $R^2$ of up to 91%. However, as noted by the authors, these $R^2$ values were overestimated and deemed unreliable due to the following reasons:

- The training data was derived exclusively from a single breed (Guernsey) due to limited data availability. This approach overlooks significant variations in HL and type traits across different breeds, limiting the model's generalizability.
- Although the dataset included a large number of type traits, the number of cows was insufficient to adequately capture the variations in these traits. This imbalance likely contributed to the inflation of the $R^2$ values.
- Since the data was collected from a single farm, the study did not account for variations in the relationships between HL and type traits across different management systems, further limiting the model's applicability.

More recently, by emerging sensing devices in modern farms, a few studies have used sensor data for lifespan predictions in dairy farms (Adriaens et al., 2020), (Ouweltjes et al., 2021), (Ranzato et al., 2022). For example, (Adriaens et al., 2020) used activity sensor data as well as daily milk production data to predict PL. They began by calculating a lifetime resilience score (RS) for each cow, derived from variables such as the last lactation number, average calving interval, age at first calving, milk production, and days in milk (DIM). Two linear regression models were then developed to predict RS, each using different feature sets. The first model relied solely on milk production features and achieved an adjusted determination coefficient ($R^2adj$) ranging from 3% to 61% across different farms. A second model, incorporating both activity sensor data and milk production features, was developed for farms equipped with activity sensors, resulting in improved $R^2adj$ values between 20% and 76%. However, this approach was limited to modern farms with access to such sensors, reducing its applicability for less technologically advanced farms. Furthermore, as highlighted by the authors, the significant variability in model performance across farms indicated that a single, generalized model would not perform well universally.

Similar to (Adriaens et al., 2020), Ouweltjes et al. in (Ouweltjes et al., 2021) also calculated and predicted the lifetime resilience score (RS) using the same approach. However, instead of linear algorithms, they employed nonlinear machine learning (ML) algorithms, including Random Forest (RF) and logistic regression (LR), to build the prediction models. Despite using more advanced algorithms, their models exhibited similar performance to (Adriaens et al., 2020). Furthermore, while (Adriaens et al., 2020) used data from 27 different farms, (Ouweltjes et al., 2021) relied on data from only one farm, limiting the generalizability of their findings.

Authors in (Ranzato et al., 2022) used sensor data along with birth, calving, and culling dates to predict dairy cows' survival at different lactations. The sensor data included daily milk yield (MY), body weight (BW), and rumination time (RUM) recorded over various stages of the first lactation for each cow. They developed a joint model that combined a mixed-effect linear model (MELM) to estimate the sensor data (MY, BW, RUM), and a Cox model to predict survival probabilities based on the estimated sensor data across different lactations. The final joint model demonstrated the ability to predict survival probabilities, achieving an average area under the curve (AUC) ranging from 45% to 76% across the studied farms.

Recent advancements in computer science have established artificial intelligence (AI), particularly deep learning (DL) techniques, as powerful and effective methods for developing decision-support tools across a wide range of fields. n dairy farming, deep learning (DL) has been predominantly applied in two major areas: health monitoring and cow identification (Mahmud et al., 2021). Additionally, DL has been utilized for predicting milk production, offering promising results in enhancing farm management and productivity (Liseune et al., 2021). However, to the best of our knowledge, no study in the literature has employed transformer-based techniques for estimating lifespan in dairy cows. Moreover, most existing approaches are constrained by the data used for building prediction models, often limited to specific age ranges, breeds, or individual farms. This highlights the absence of a generalizable model applicable to diverse farming conditions, including variations in cow age and breed. In contrast, our study utilized data collected from 7 different farms, encompassing 221 unique breed combinations and cows of varying ages, ranging from a few months to over 120 months old. Therefore, this study stands out due to its emphasis on the following aspects:

- Innovative Methodology: This work is the first to explore transformer-based deep learning techniques for estimating herd life (HL) in dairy cows.
- Comprehensive Dataset: The model was developed using data collected from 7 farms, encompassing 221 breed combinations and cows of varying ages, ranging from a few months to over 120 months. This ensures the model's broad applicability across diverse farming conditions and cow populations.
- Improved Performance: The proposed model achieves at least 10% better overall performance in HL estimation compared to existing methods, providing a more accurate and generalizable solution for decision support in dairy farming.

**Table I**. Summary of data-driven approaches used for herd life estimations in dairy cows (DIM: days in milk, MIM: months in milk, RS: resilience score, AUC: area under curve, RF: Random Forest, LogReg: Logistic Regression, MELM: Mixed-effect linear model, RUM: rumination time)

| Reference | Key Features | Outcomes to Predict | No. of records | Modelling Approach | Model Performance |
|---|---|---|---|---|---|
| (Honnette et al., 1980) | Final score, First lactation milk, Ages at 1st calving, DIM, type traits | Herd life, lifetime milk | ~35K Holstein cows | Linear Regression | - $R^2$ = up to 11.2% for predicting herd life<br>- $R^2$ = up to 15.2% for predicting lifetime milk |
| (VanRaden & Klaaskate, 1993) | Cumulative MIM, current MIM, age at first lactation, current months dry, lactation status | productive life | ~2.6M Holstein cows | Linear Regression | $R^2$ = 8% to 59% |
| (Cruickshank et al., 2002) | combination of production and genetic type traits | Herd life | ~19K Guernsey cows | Mixed-effect linear model | - $R^2$ when using only type traits = up to 50%<br>- $R^2$ when using type and yield traits = up to 91% (1 farm only) |
| (Adriaens et al., 2020) | Activity sensor data, , last lactation number, average calving interval, age at first calving, milk production and DIM | RS | 27 farms with ~4K unique cows | Linear Regression | - $R^2_{adj}$ for the farms without sensor data = 3% to 61%<br>- $R^2_{adj}$ for the farms with sensor data = 20% to 76% |
| (Ouweltjes et al., 2021) | Birth date, insemination date, calving date, culling date, health treatments and sensor data including milk yield, body weight, activity, and rumination. | RS | 1 farm with ~2K cows | Joint model included:<br>- RF model to classify cows based on RS categories.<br>- LogReg model for feature selection | Accuracy = ~50% overall |
| (Ranzato et al., 2022) | sensor data as well as birth date, calving date, culling date and sensor data including daily milk yield, body weight, and RUM | Cow survival at different lactations | 6 farms with ~1K Holstein cows | Joint model included:<br>- MELM for estimating the sensor data<br>- Cox model to predict the survival probabilities | AUC = 45% to 76% across the farms. |

## 2 Materials and Methods

### 2.1 Data Processing

Figure 1 shows an overview of the data processing pipeline (DPP) used in this work. The initial data processing stage involved combining different datasets, as detailed in Table II. This way, we can leverage all the diverse variables and traits available across the datasets. The primary key for combining the datasets is national ID (NID), which is a unique identifier for each cow. The resulting data is then cleansed to rectify typos, identify and handle outliers, and remove unnecessary columns such as NID and herd ID. We have further dropped variables exhibiting high collinearity to ensure more robust and reliable results. Then, we transform the data based on the following rules:

- For each cow, we calculate the HL in days from birth to culling dates.
- For each cow, 3 events have been recorded in the data throughout her lifetime including test, breeding and treatment. To analyse the impact of these events on the HL of each cow, we create a binary variable for each event. For example, if a cow is treated on a particular date, the corresponding value is set to 1. Otherwise, it is set to 0.
- The DPP produces 2 outputs including tabular and sequential data. Tabular data is used to build models other than the TM. However, as TMs are inherently sequence-based models, they exclusively process data in the form of sequences (Lin et al., 2022). Therefore, it is necessary to convert the tabular data obtained from the previous steps into sequential data. This conversion will yield a large array of multivariate sequences, with each sequence corresponding to an individual cow.

**Table II**. List of datasets used along with sample variables (ABV: Australian Breeding Value)

| Dataset | Sample Variables |
|---|---|
| DS102: Cow Pedigree Record | 'National Cow ID', 'National Herd ID', 'Within-Herd Cow ID', 'Birth Date', 'Sire National ID', 'Dam National ID', 'Animal Termination Code', 'Animal Termination Date' |
| DS103: Lactation Record | 'Milk Yield', 'Fat Yield', 'Total Solids 305', 'Milk 305', 'Fat 305', 'Protein 305', 'Protein Yield', 'Lactose Yield', 'Solids Yield', 'PI Milk', 'PI Fat', 'PI Protein', 'Custom PI', 'National Cow ID', 'National Herd ID', 'Within-Herd Cow ID', 'Calving Date', 'Calving Code', 'Parity', 'Termination Date', 'Termination Code', 'Num PI TEST', 'Lactose 305' |
| DS104: Test Day Record | 'National Cow ID', 'National Herd ID', 'Within-Herd Cow ID', 'Test Date', 'Fat Percentage', 'Protein Percentage', 'Lactose Percentage', 'Somatic Cell Count', 'Milk Yield', 'Calving Date' |
| DS108: Pregnancy Test Record | 'National Cow Id', 'National Herd Id', 'Within-Herd Cow Id', 'Date', 'Code', 'Result', 'Bull National Id', 'Technician Code' |
| DS112: Calving Ease Record | 'National Cow ID', 'National Herd ID', 'Within-Herd Cow ID', 'Calving Date', 'Parity', 'Last Mating Date', 'Litter Size', 'Calving Ease', 'Sex Of Calf', 'Fate Of Calf', 'Size Of Calf' |
| DS116: Herd Health Record | 'National Cow ID', 'National Herd ID', 'Date', 'Health Event Code', 'Health Treatment Code', 'Anatomical Position' |
| DS202: ABV | 'National ID', 'National Herd ID', 'Within-Herd Cow ID', 'Breed Of Cow', 'Date Of Birth', 'Mammary System', 'Health Weighted Index', 'ABV Mastitis Resistance', 'Reliability Mastitis Resistance' |

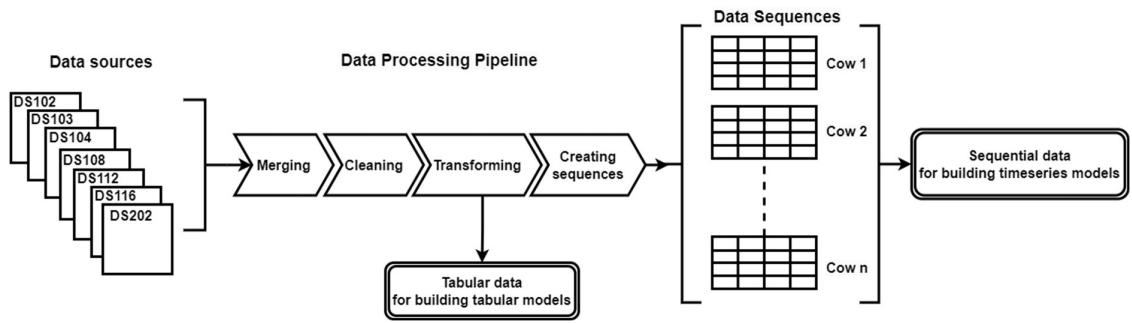

**Figure 1.** Overview of data processing pipeline in the study.

## 2.2 Model Building

As shown in Figure 1, the DPP returns data in 2 different formats including tabular and sequential data. The sequential data is then used for training a TM. However, for the purpose of comparison, the tabular data was used for training other models including linear regression, General Linear Model (GLM) and RF. Due to the superiority of transformers in modelling long dependencies in sequential data (Ahmed et al., 2023), the TM showed the best performance compared to the other models. The comparative results are shown in **Error! Reference source not found.** in Section 3.

## 3 Results and discussion

### 3.1 Input and Output Variables

As shown in Table III, we used multiple datasets with totally ~782K records across different farms across Australia. The data was related to ~19K unique cows from 221 various breeds. Table III also illustrates list of variables used for building the models. Out of totally 50 fields in the merged dataset, 33 variables have been removed during the data processing step including ID fields, variables with high missing rate and high collinearity. Table IV shows the statistics of the selected variables with numerical type.

**Table III**. Summary of variables used in this work. (PI: Production index, SCC: somatic cell counts, HWI: health weighted index, ABV: Australian breeding value)

| Attribute Name | Outcomes to Predict | Data used in this study |
|---|---|---|
| Current life (days)<br>Lactation<br>Milk 305<br>Lactose Yield<br>PI Fat<br>Number of Tests in PI<br>Milk<br>Fat<br>Lactose Percentage<br>SCC<br>Days Pregnant<br>Mammary System<br>HWI<br>ABV Mastitis Resistance<br>Tested (0/1)<br>Breed (0/1)<br>Treated (0/1) | • Classification model: herd life in days<br>• Regression model: herd life categories including high, medium and low. | The data was collected from 7 different farms across Australia and included ~782K records related to ~19K unique cows |

**Table IV**. Statistics of numerical variables used in modeling

| Trait | Mean | STD | Min | Max |
|---|---|---|---|---|
| Lactation | 3 | 1.8 | 0 | 12 |
| Milk | 23.4 | 8.6 | 0.1 | 77.2 |
| Fat | 4.2 | 0.8 | 0 | 11.5 |
| Lactose | 4.6 | 1.2 | 0 | 6.5 |
| Milk 305 | 6847 | 2241 | 0 | 17910 |
| Lactose Yield | 331 | 237 | 0 | 2966 |
| PI Fat | 95 | 25 | 0 | 168 |
| No. of PI Test | 7.5 | 4 | 0 | 39 |
| SCC | 162.5 | 437.6 | 1 | 13125 |
| HWI | 35.3 | 93 | -342 | 454 |
| Mamary System | 94 | 6 | 68 | 114 |
| Mastitis Resistance | 101 | 2.7 | 87 | 110 |
| HL | 2617 | 898 | 101 | 4993 |

Total HL of cows used in this work ranged mostly between 100 and 5000 days. A few cows had HL out of this range, which was removed from the dataset as outliers. Figure 2 shows how HL is distributed across cows in the dataset.

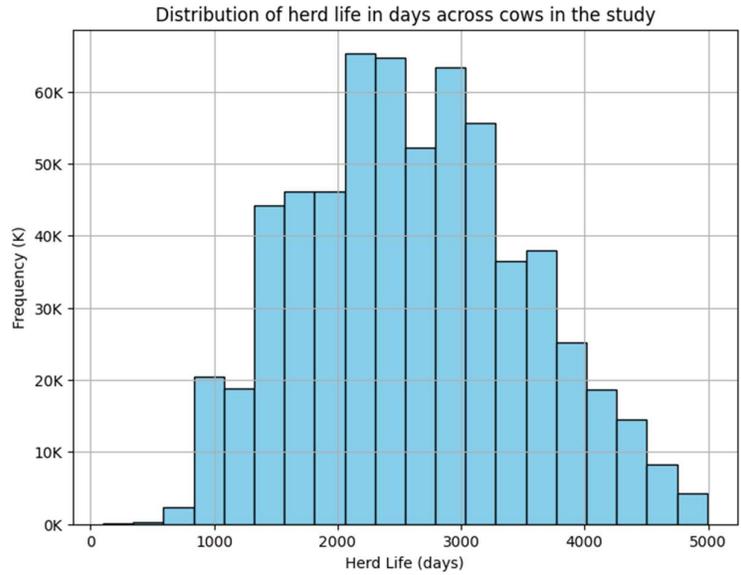

**Figure 2.** Distribution of herd life of cows used in the study.

## 3.2  Correlation Analysis

Figure 3 shows the linear correlation between HL and the selected variables excluding current life (CL) in days. The linear correlation between CL and HL is approximately 60%, which is evident. Therefore, we have omitted CL from Figure 3 to enhance focus on the remaining variables. Among the selected variables, Lactose and Lactation have the highest negative and positive linear correlation with HL, respectively. The low values in Figure 3 indicate a weak linear correlation between HL and the selected variables, excluding CL. This observation further confirms why the linear methods utilized in the literature for estimating HL in dairy cows often exhibit poor performances.

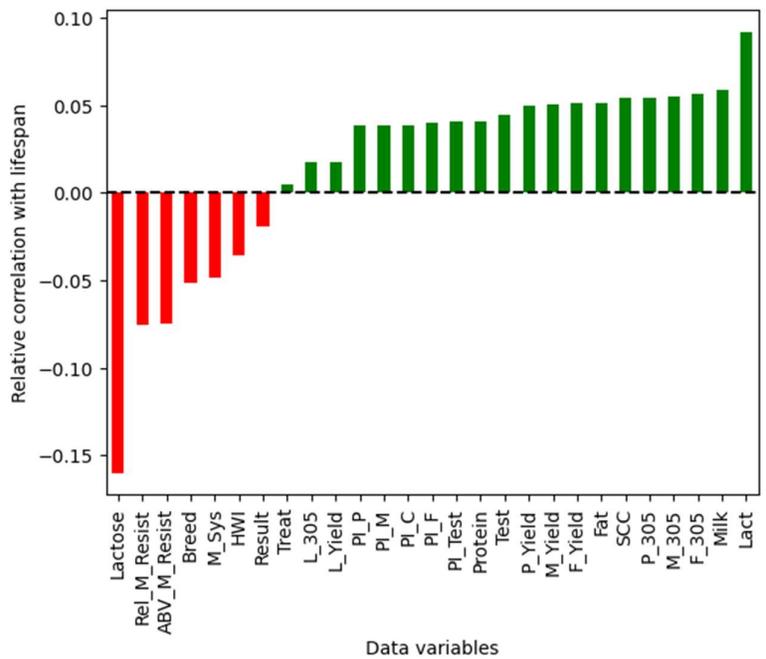

**Figure 3**. Linear correlation between herd life and attributes used in the study.

In addition to the linear methods, we have also used a tree-based algorithm as a conventional nonlinear method. Figure 4 shows the relative importance of the selected variables (excluding CL) with HL when using an RF model. Following CL with an approximate relative importance value (RIV) of 0.59, fat with only 0.12 RIV is the most significant features when estimating HL using an RF model. This highlights the limited contribution of variables other than CL in the estimation of HL, even with conventional nonlinear models such as RF.

### 3.3 Sequence Extraction

Transformers require sequences with equal lengths. However, in the raw data, the number of records per sequence varies, as different numbers of events are recorded for each cow. Therefore, before inputting data into a TM, sequences must be tailored to ensure that all sequences have an identical number of records, denoted as $L$. In this process, longer sequences with more than $L$ records should be truncated, while sequences shorter than $L$ should be padded with zeros. The length of sequences ($L$) is a hyperparameter and finding the optimal value requires to train models with different sequence lengths. Additionally, selecting the appropriate sequence length depends on the number of records available per cow in the dataset. Figure 5 shows the distribution of number of records per cow in the dataset. The number of records per cow in the dataset ranges from 1 to 200. On average, there are around 40 records per each cow in the dataset.

To find the optimal sequence length, we trained multiple models using sequences with different lengths including 5, 10, 20 and 40. The best results achieved by the model trained by sequences with $L=10$. As shown in Figure 6, the model trained by sequences with $L=10$ (*model_seq10*), can approximately achieve an $R^2$ value of 82% when using the 10 latest records per each cow. It also shows that by even using only the 5 latest records per each cow, the *model_seq10* can achieve nearly 79% as $R^2$. This is more interesting as over 92% of cows in the dataset have more than 5 records as shown in Figure 5.b.

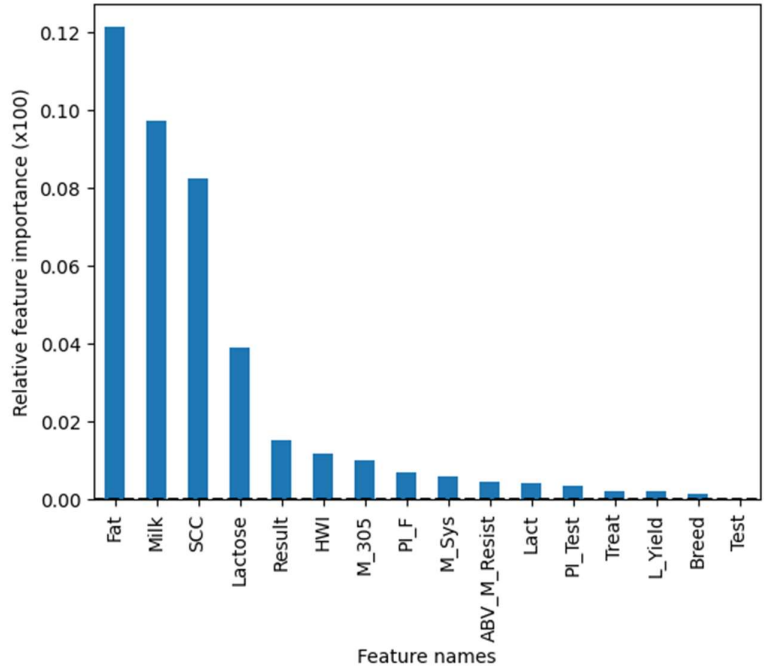

**Figure 4**. Relative importance of selected variables with HL using random forest model

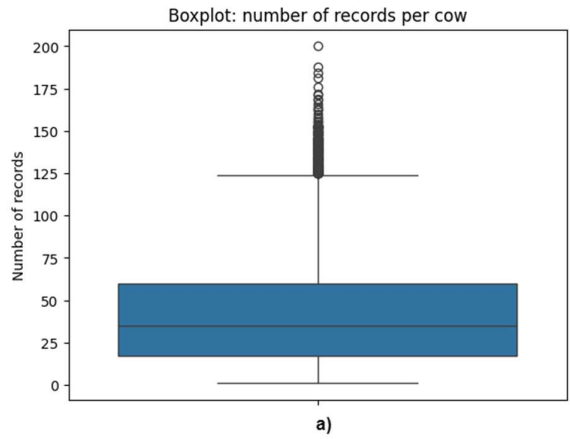

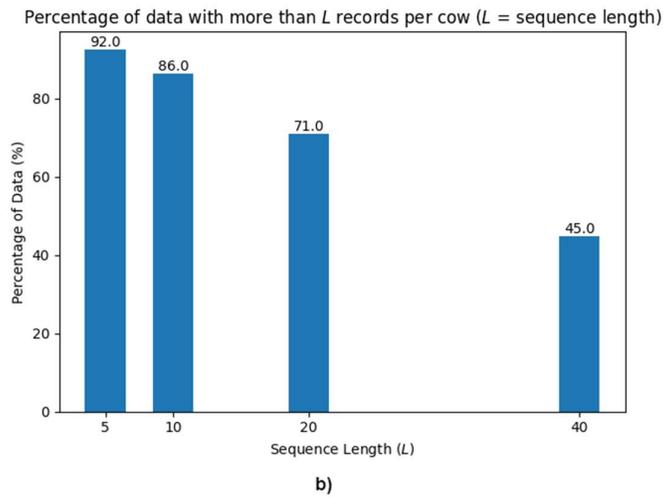

**Figure 5**. Distribution of number of records per cow in the dataset: a) boxplot, b) data percentage

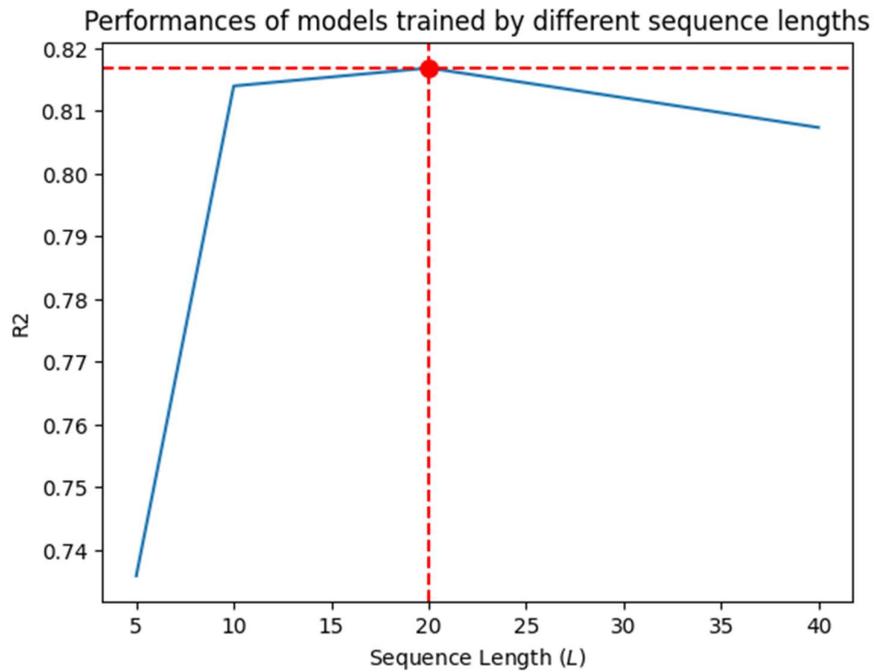

**Figure 6**. Performances of models trained by different sequence lengths

## 3.4 Model Evaluation

Out of ~870K total records in the dataset, we used 80% for training and 20% for testing purposes. Like the relevant works showed in Table I, we have built multiple models for comparison purposes for both regression and classification approaches. Models used in the regression approach included GLM, MELM, and linear regression model (LRM). Additionally in comparison with the RF-based classifier used in (Ouweltjes et al., 2021), we have also demonstrated the performance of the TM as a classifier for categorizing HL into high, medium and low. Table V presents the performance of the proposed TM in comparison with the models utilized in other studies for both regression and classification approaches. Using the same data with a similar processing method, the TM used in this work showed an $R^2$ value of 82% and an accuracy of 85% for regression and classification, respectively.

Table V. Comparing prediction performance across different models within each method.

| Methods | Models | Prediction Performance |
|---|---|---|
| Regression | MELM | $R^2 = 0.23$ |
|  | Linear Regression | $R^2 = 0.37$ |
|  | GLM | $R^2 = 0.38$ |
|  | Transformer (this work) | $\boldsymbol{R^2 = 0.82}$ |
| Classification | RF | $Accuracy = 55\%$ |
|  | Transformer (this work) | $\boldsymbol{Accuracy = 83\%}$ |

Figure 7 and Table VI show how the TM performs when classifying HL into high, medium and low. As used in (Ouweltjes et al., 2021), the misclassification between cows with low or high HL is a key metric in evaluating a model performance and is called critically misclassified (CritMis) cows. As illustrated, none of the cows with actual low HL has been predicted as high that means zero CritMis for cows with low HL. Additionally, the CritMis for cows with high HL is only 34 out of 674 that is equal to 5%. The overall accuracy of the classifier model across all the classes is 85%.

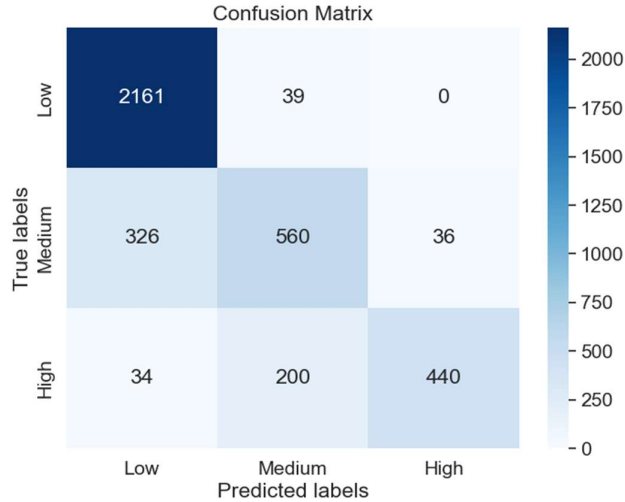

**Figure 7**. Confusion matrix for transformer model when classifying lifespan into low (<2158 days), MEDIUM (between 2158 and 2997 days), and high (>2997 days).

**Table VI**. Performance report for transformer model when classifying lifespan into low (<2158 days), MEDIUM (between 2158 and 2997 days), and high (>2997 days).

| Class | Precision (%) | Recall (%) | F1-score (%) | Support (%) |
|---|---|---|---|---|
| Low | 86 | 98 | 92 | 2200 |
| Medium | 70 | 61 | 65 | 922 |
| High | 92 | 65 | 77 | 674 |
| Overall Accuracy (%) | 83 |  |  | 3796 |

Table VII shows herd-level performance metrics across different farms in the study. As illustrated, the range of $R^2$, MAE and accuracy across different farms are 65%-88%, 211-345 days and 78%-91%, respectively.

Table VII. Performances of proposed regression and classification models across various farms

| Herd No. | Regression | | Classification |
|---|---|---|---|
| | R2 (%) | MAE (days) | Accuracy (%) |
| 240122D | 86 | 251 | 89 |
| 2B0057O | 82 | 284 | 78 |
| 2K0054J | 65 | 345 | 85 |
| 4A1819R | 81 | 286 | 80 |
| 540162K | 84 | 227 | 83 |
| 540435C | 85 | 254 | 81 |
| 8K0078G | 88 | 211 | 91 |
| Overall | 82 | 266 | 83 |

## 4 Conclusion

In this study, we developed 6 different models for predicting herd life in dairy cows using available historical data in farms. Extensive testing on 870K records from 19K unique cows across 7 different farms across Australia demonstrated that transformer-based models outperformed other models, achieving an overall $R^2$ of 82% in regression tasks and an overall accuracy of 83% in classification approaches. Additionally, models built based on transformers showed greater robustness across various farms compared to other models.

The results of this study highlight the potential of using transformers as the latest AI technology in modern data-driven dairy farm management practices. This could pave the way for the application of even more advanced AI technologies such as generative AI models, which are also built based on transformers, in dairy farm management.

**CRediT authorship contribution statement**
**Mahdi Saki:** Writing – review & editing, Writing – original draft, Investigation, Data curation, Methodology, Software, Conceptualization, Visualization, Validation. **Justin Lipman:** Writing – review & editing, Supervision, Project administration.

**Declaration of competing interest**
The authors declare that they have no known competing financial interests or personal relationships that could have appeared to influence the work reported in this paper.

**Acknowledgments**
This study was conducted as part of Food Agility project FA 078 – On-farm mastitis, jointly funded by Food Agility CRC, Dairy Australia, Coles Sustainable Dairy Development Group, Charles Sturt University, University of Technology Sydney, and the University of Sydney. We would like to thank Dr. Stephanie Bullen from Dairy Australia for her insightful comments and quick review of an early draft of this paper.

**Data availability**
Data for this study was provided from Australian dairy herds and Ginfo farmers who participated in milk recording and collecting other on-farm data from a database that is maintained by DataGene Pty Ltd., (Melbourne, Australia). The data of this project cannot be shared due to commercial and privacy policies.